\documentclass[12pt]{spieman}  
\usepackage{subcaption}
\usepackage{amsmath,amsfonts,amssymb}
\usepackage{graphicx}
\usepackage{setspace}
\usepackage{csquotes}
\usepackage{hyperref}

\title{On the Construction of Distribution-Free Prediction Intervals for an
Image Regression Problem in Semiconductor Manufacturing}

\author{Inimfon I. Akpabio}
\author{Serap A. Savari}
\affil{Texas A\&M University, Mail Stop 3128 TAMU, College Station, TX 77843-3128, USA}

\begin{document}

\maketitle

\begin{abstract}
The high-volume manufacturing of the next generation of semiconductor devices 
requires advances in measurement signal analysis. Many in the semiconductor 
manufacturing community have reservations about the adoption of deep learning;
they instead prefer other model-based approaches for some image regression 
problems, and according to the 2021 IEEE International Roadmap for Devices and
Systems (IRDS) report on Metrology a SEMI standardization committee may 
endorse this philosophy.  The semiconductor manufacturing community does, 
however, communicate a need for state-of-the-art statistical analyses to 
reduce measurement uncertainty.  Prediction intervals which characterize 
the reliability of the predictive performance of regression models can 
impact decisions, build trust in machine learning, and be applied to other 
regression models.  However, we are not aware of effective and sufficiently
simple distribution-free approaches that offer valid coverage for
important classes of image data, so we consider the distribution-free
conformal prediction and conformalized quantile regression framework.
The image regression problem that is the focus of this paper pertains to 
line edge roughness (LER) estimation from noisy scanning electron microscopy 
images.  LER affects semiconductor device performance and reliability as well 
as the yield of the manufacturing process; the 2021 IRDS emphasizes the 
crucial importance of LER by devoting a white paper to it in addition to 
mentioning or discussing it in the reports of multiple international focus 
teams.  It is not immediately apparent how to effectively use normalized 
conformal prediction and quantile regression for LER estimation.
The modeling techniques we apply appear to be novel for finding 
distribution-free prediction intervals for image data and will be 
presented at the 2022 SEMI Advanced Semiconductor Manufacturing Conference.
\end{abstract}

{\noindent \footnotesize\textbf{*}Inimfon Akpabio,  
\linkable{nini16@tamu.edu}; Serap Savari, \linkable{savari@tamu.edu} }

\begin{spacing}{1}   

\section{Introduction}
The members of the semiconductor manufacturing community have enabled
today's technology.  They are the unsung heroes of the machine learning
revolution, and they will impact the future of computing.  Their work is 
complex, and we mention only one aspect of it here.  There is a lengthy 
and intricate process to fabricate integrated circuits on the surface of a 
silicon crystal wafer.  That process requires hundreds of inspection and 
measurement steps \cite{hitachi}, and metrology is the ``science of 
measurement, embracing both experimental and theoretical determinations at 
any level of uncertainty in any field of science and technology'' 
\cite[p.~1]{ssv2015}.  The semiconductor industry has the most stringent 
metrology requirements in the entire manufacturing sector 
\cite[p.~xii]{ssv2015} because the billions of devices in a chip must
all operate to a tight specification \cite{orji}.  Much of the measurement
data is in the form of digital images, and there is a continual 
need to improve the process information from microscopy techniques and
measurements as technology evolves \cite[\S 5.3]{irds-met}.  More generally,
the future of semiconductor device fabrication will have growing needs for
data processing, information extraction, and knowledge management
\cite{irds-met, irds-fi, irds-ye}.  Therefore, machine learning may be
able to address some of these needs, and an April~2021 
McKinsey \& Company article attempts to persuade semiconductor-device 
makers to better capture the significant value-creation potential of 
artificial intelligence \cite{mckinsey2021}.
Yet in semiconductor metrology, ``the promise of advanced data analytics 
(in all its variants) has not been realized'' \cite[\S 9]{irds-met}.  
The Call for Abstracts for this year's SEMI Advanced Semiconductor 
Manufacturing Conference (ASMC 2022) lists the ``pros and cons of machine 
learning in measurement signal analysis'' \cite{call} as a topic of interest.  
So what are some of the fundamental issues?

Human factors are affecting the digital transformation of semiconductor
manufacturing and other industries \cite{keil, saldanha}.  
Regarding machine learning, there is another widespread problem.  
In December 2020 the National Science Foundation and the National Institute 
of Standards and Technology hosted a gathering of many leaders throughout the 
manufacturing sector to consider the acceleration of the implementation of 
artificial intelligence in manufacturing.  The complaints of real and 
perceived risks and a lack of transparency are two of the obstacles to the 
wider adoption of artificial intelligence \cite{nsf-nist, b7}, and in a
different discipline there have been concerns expressed about it in 
connection with magnetic resonance imaging\cite{mri}.  
Moreover, within semiconductor manufacturing the desire to better understand 
the uncertainties associated with prediction engines 
extends beyond machine learning models.  Section~5.8.4.2 of the 2021 IEEE 
International Roadmap for Devices and Systems (IRDS) report on Factory 
Integration solicits the incorporation of indications of the quality of 
predictions in its prediction vision \cite{irds-fi}.  Section~4.2.10 of the
2021 IRDS Executive Summary seeks state-of-the-art statistical analyses
to assist in decreasing the measurement uncertainty of sub-7~nm process
control \cite{irds-es}.

One way to address these issues is through uncertainty quantification
(see, e.g., References~15 and 16), and a relatively 
simple approach is the use of prediction intervals which characterize 
the reliability of predictive performance.  For a regression model outputting 
a single number a prediction interval offers a range of values in which the 
output variable lies with high probability.  The design of prediction 
intervals has been extensively studied and has certain objectives.  First,
the underlying modeling assumptions should be well-suited to the 
application in order to provide valid coverage.  Second, they should produce
the shortest possible intervals to help decision makers.
Finally, they need an appropriate computational complexity
\cite[p.~41]{irds-fi}.

Conformal prediction \cite{vgs2005, sv2008, ph2011} is a simple, mathematically
rigorous, and distribution-free guideline to design prediction intervals;
one new variant \cite{rpc19, kjl20} combines it with a widely used technique
known as quantile regression \cite{kb78}.  This framework assumes the
{\em exchangeability} of data, which roughly means that the examples are
representative and their ordering does not matter.
In a recent publication\cite{b6} we introduced the nanofabrication
community to conformal prediction and considered the image
regression problem of line edge roughness (LER) estimation; LER is 
important enough in semiconductor manufacturing 
to warrant a 2021 IRDS white paper \cite{irds-ler}.
While quantile regression and conformal prediction are widely used for
the construction of prediction intervals, it is not apparent how to 
effectively apply them to image data, and because of the high dimensionality
of image data this question appears to be related to the topic of 
meta-learning \cite{meta} for this application.  
While our focus is on an LER 
estimation problem, it may also be interesting to construct prediction 
intervals for some medical imaging problems such as skeletal bone age 
assessment from X-ray images \cite{xray}.  In this paper we present for 
a data analytics audience the work of an upcoming conference paper 
\cite{asmc2022} (see also Ref.~28) together with some background 
material.  In Section~2,
we provide more information about the topic of LER.  In Section~3, we
briefly review conformal prediction and quantile regression.  In 
Section~4, we discuss the deep convolutional neural network
EDGENet \cite{bacus2018, csy2019}, which is the image regression model
we study.  In Section~5, we discuss the new neural networks we introduce
to assist with the construction of prediction intervals; these are founded
on denoising and successes from image recognition.  In Section~6,
we provide experimental results.  In Section~7, we conclude the paper.

\section{Background on LER}
For many years, the semiconductor manufacturing industry mainly relied upon
photolithography or optical lithography to fabricate multiple integrated
circuits on silicon wafers. Here, an image is formed when 
a thin-film material is selectively exposed to light at 193~nm wavelength
\cite{Lin}.  The newer optical technology extreme ultraviolet (EUV) lithography
is now in use for the most advanced resolution requirements.
In older technologies with larger feature sizes, a mean field theory
approximation is adequate to model the effects of the exposure process
\cite{mack2014}.
However, as feature sizes shrink this approximation no longer suffices to
capture the randomness inherent in the lithographic process, and this
randomness partially manifests itself as roughness or fluctuations in the
edges and widths of features.  This roughness has not been scaling at the
same rate as feature sizes.  This form of uncertainty is increasingly
important in linewidth control and is impacting the ultimate limits of
resolution in semiconductor lithography \cite{mack2014}.
The newer EUV technology uses light at 13.5~nm wavelength to produce
smaller circuit features, but statistical fluctuations are growing more
important as dimensions shrink \cite[\S 2.2.9]{irds-es}.

In order for integrated circuit manufacturing to be economically viable,
the {\em yield} or the fraction of the produced product with adequate quality
must be adequate.  To avoid adverse consequences to delay and yield,
it is necessary for entire transistors to have uniform characteristics,
so it is necessary to estimate and manage random variations on 
transistor performance including LER \cite[pp.~12-14]{shin}.
LER refers to the edge displacements of a feature edge from a mean edge 
position and is known to be a crucial factor in the yield of integrated 
circuit manufacturing (see, e.g., \cite[pp.~82-92]{hjl}).
Nanostructure geometry reconstruction algorithms 
generally require edge-contour extraction algorithms whose performances
depend on imaging technologies.

The critical-dimension scanning electron microscope (CD-SEM) is the standard
metrology tool to obtain measurements to analyze LER 
\cite[pp.~109, 114]{ssv2015}.  Here, a focused beam of electrons scans and
interacts with the surface of a sample to produce an image of the sample at 
sub-micron resolutions which contain information about the sample surface
geometry.  The image quality depends on the energy of the electron
beam or the dose provided.  High-dose CD-SEM images cause sample damage 
during repeated measurements \cite[p.~42]{ssv2015} and have higher acquisition
times.  Low-dose CD-SEM images have more artifacts in the form of noise, 
blur, edge effects and other instrument errors which hinder accurate 
estimation, but they are desirable for manufacturing since they reduce 
sample damage and acquisition time (see, e.g., 
Ref.~35).  Moreover, the near-term ``grand challenges'' the 2021 IRDS 
Executive Summary specifies for yield enhancement include 
\cite[\S 4.1.11.1]{irds-es}
\begin{itemize}
\item ``Existing techniques trade off throughput for sensitivity, but at
expected defect levels, both throughput and sensitivity are necessary
for statistical validity.''
\item ``Reduction of inspection costs and increase of throughput is crucial
in view of cost of ownership.''
\item ``Detection of line roughness due to process variation.''
\item ``Reduction of background noise from detection units and samples to
improve the sensitivity of systems.''
\item ``Improvement of signal to noise ratio to delineate defect from
process variation.''
\item ``Where does process variation stop and defect start?''
\end{itemize}
Since deep learning \cite{nature} has altered the practices of signal, image, 
and video processing, its rapid advances may offer the potential to better
address these issues, 
particularly since line and contour measurements currently require an
average of multiple lines or contours in order to sufficiently improve
precision (see, e.g., \cite[\S 10.6]{ssv2015} and \cite[p.~11]{irds-met}).
There are some efforts in the industry to use deep learning for defect and
fault detection in the fabrication process.  Nevertheless, the semiconductor 
metrology community began to handle these types of
problems long before the advent of deep learning, and model-based metrology
is respected \cite[p.~11]{irds-met}.  Yet any model is limited by its
assumptions.  According to \cite[\S 5.4.1]{irds-met}, a SEMI Standards
Committee is updating the determination of LER estimation to incorporate
a specific model-based methodology. That section of the document states:
\begin{quote}
``Another important factor in measurement of ... LER on imaging tools is 
edge detection noise. ... A methodology has been demonstrated to remove
this noise term, leading to an unbiased estimation of the roughness.
Use of this is deemed very important to ensuring accuracy of roughness
measurements in the future and should be a key ingredient in allowing for
inter-comparison of data across the litho-metrology community.''
\end{quote}
In statistics, unbiased estimation means that the expected 
value of the estimate equals the true value of the parameter being estimated 
(see, e.g., Ref.~37).  There continue to be statistical studies on how
CD-SEM metrology artifacts affect the measurement of roughness and on
monitoring CD-SEM tool health\cite{b9}, so not all aspects of CD-SEM operation
for LER estimation are equally well understood.
Furthermore, the variance of the estimation error is often of interest to 
statisticians (see, e.g., Ref.~37), and this topic is not discussed in 
\cite[\S 5.4.1]{irds-met}.
Moreover, while the inter-comparison of data is a worthy objective, we are not
aware of published comparisons of the model-based methodology with any 
deep learning-based LER estimation algorithm.  A comparison of different 
approaches and algorithms is, in principle, possible.  
Simulation has for decades been a criical and necessary tool to the success 
and advancement of semiconductor manufacturing \cite{mack2005}, and it is 
projected to play a larger role going forward including its integration into 
all aspects of metrology \cite[p.~3]{irds-met}.  As we will see, one can use 
simulation to generate datasets to create and test algorithms.  Moreover, 
conformal prediction is popular in part because it can offer a diagnostic tool 
for and a comparison tool among regression models \cite{lei2018, sc20}.  
LER evaluation methods 
continue to be an active area of research \cite[\S 5.4.1]{irds-met}.  

Our investigation considers a basic version of an LER estimation problem
in which the dataset consists of a collection of $64 \times 1024$ simulated
SEM images each containing one rough line and an unknown level 
of Poisson noise.  Our group's deep convolutional neural network (CNN) \\
EDGENet \cite{bacus2018, csy2019} directly outputs a matrix of dimension 
$2 \times 1024$ with the estimated left and right edge positions of the line; 
we will discuss EDGENet and the simulated dataset used to study it, and
we point the audience to References~4 and 24 for a sample of the extensive
literature on LER and its estimation.  Since CD-SEM metrology artifacts
affect the accuracy of LER measurements\cite{b9}, we propose denoising
as a first step in constructing prediction intervals and apply our group's
Poisson denoising CNN SEMNet \cite{emlc2018, csy2019}, 
which was designed for the 
same dataset as EDGENet.  We use various computer vision and image processing 
techniques in combination with the conformal prediction and conformalized 
quantile regression frameworks to examine how the EDGENet LER prediction errors
are related to the ``noise image'' defined as the absolute difference
between a noisy input image and its associated denoised output image from
SEMNet.

\section{On Conformal Prediction and Conformalized Quantile Regression}

Since simulation is a useful and accepted tool for semiconductor metrology,
we have the flexibility to use supervised machine learning on collections 
of input/output data pairs $z^i = (x^i, \ y^i)$, where the input $x^i$ refers
to a noisy SEM image and the output $y^i$ 
is a scalar.  We assume that the training instances 
$z^1, \ z^2, \ \dots , \ z^n$ are exchangeable and $\alpha \in (0, \ 1)$ 
is a prespecified miscoverage rate.  For a new test input $X^{n+1}$ our goal
is to construct a {\em marginal} prediction interval $C(X^{n+1})$ such that
\begin{displaymath}
P[ Y^{n+1} \in C(X^{n+1})] \geq 1 - \alpha.
\end{displaymath}
Marginal coverage occurs on average and is weaker than conditional coverage
on a specific observation $X^{n+1}=x^{n+1}$ \cite{lei2018, sc20}.  
Nevertheless, it is advantageous that the framework does not require
distributional assumptions on the data beyond exchangeability.

We apply the split or inductive conformal predictive methodology\cite{vgs2005}
and partition the input/output data pairs into a proper training set 
${\cal Z}_T$, which is used to train a regression model we will call $g$, 
a calibration set ${\cal Z}_C$, which is processed using a designated
``nonconformity score'' to offer a way to
construct prediction intervals for future examples, and a test set
${\cal Z}_{\tau}$, which is used to evaluate the predictive model.

The ``residual'' $\eta$ for an instance $z^i = (x^i , y^i)$
and a regression model $g$ is the most basic nonconformity score:
\begin{displaymath}
\eta (z^i) = | y^i - g(x^i) |.
\end{displaymath}
Let $r_1 , \ r_2 , \ \dots , \ r_{| {\cal Z}_C |}$ be the sorted
nonconformity measures of each example in the calibration set 
in nonincreasing order and define 
$m = \lfloor \alpha (| {\cal Z}_C | + 1) \rfloor$. Then for a new
input image $X^j = x^j$ the prediction interval
\begin{displaymath}
[g(x^j) - r_m, \; g(x^j) + r_m]
\end{displaymath}
offers valid marginal coverage, but the fixed prediction interval width offers
no information about the difficulty of edge detection corresponding to $x^j$.

To account for this weakness and potentially reduce the average prediction
interval length\cite{ph2011}, we fit a model $\gamma$ to the training set
to study the relationships between a noise image $x_N^i = |x^i - x_D^i |$,
where $x_D^i$ is the output of the denoiser SEMNet \cite{emlc2018, csy2019} 
for input $x^i$, and the corresponding residual or absolute prediction error; 
this approach appears to be related to meta-learning \cite{meta} for this 
application since there is a correlation between noise and other artifacts
and the difficulty of edge detection\cite{b9}. 
The normalized nonconformity score associated with $\gamma$ is 
\begin{displaymath}
\eta_N (z^i) \; = \; \frac{ | y^i - g(x^i) |}{\gamma (x_N^i) }.
\end{displaymath}
Let $\varrho_1 , \ \varrho_2 , \ \dots , \ \varrho_{| {\cal Z}_C |}$ be the 
sorted nonconformity measures of each example in the calibration set in 
nonincreasing order.  Then for a new input image $X^j = x^j$, the 
prediction interval
\begin{displaymath}
[g(x^j) - \varrho_m \cdot \gamma (x_N^j) , \; 
g(x^j) + \varrho_m \cdot \gamma (x_N^j) ]
\end{displaymath}
provides valid marginal coverage.  The additional flexibility potentially 
offers more efficient prediction intervals, but the question is how to
select $\gamma$.  Our general approach is inspired by aspects of
\cite[Equation (16)]{ph2011} and our previous work\cite{b6}.
We propose three variants of a deep neural network model $\phi$ which
seeks to fit a noise image $x_N^i$ to a function of the corresponding
EDGENet absolute prediction error, specifically $- \ln | y^i - g(x^i)|,$
and we set $\gamma (x_N^i) = e^{- \phi (x_N^i) }$.
We will discuss these neural network architectures in Section~5.  

The original conformal prediction framework always centers a prediction
interval at the output of the regression model being analyzed.
For potentially more efficient intervals, the conformalized quantile
regression framework \cite{rpc19, kjl20} alters the classical quantile
regression methodology\cite{kb78} to offer marginal coverage guarantees under
the exchangeability assumption; it is necessary, however, to train upper and
lower conditional quantile functions, which can be obtained by inverting the 
conditional cumulative distribution function of random variable $Y$ given 
random vector $X$, i.e., \mbox{$P[Y \leq y \ | \ X = x]$}.  
For $\epsilon \in [0, \ 1]$,
\begin{displaymath}
q_{\epsilon} (x) = \inf \{ y: \ P[Y \leq y \ | \ X = x] \geq \epsilon \}.
\end{displaymath}
For a prespecified miscoverage rate $\alpha$, there are many ways to
choose the upper and lower quantiles; we consider $\epsilon = 0.5 \alpha$
for the lower quantile and $\epsilon = 1 - 0.5 \alpha$ for the upper quantile
and estimate these from training data with the pinball loss function
\cite{kb78}:
\begin{displaymath}
\rho_{\epsilon} ( y, \ \hat{y}) 
= \left\{ \begin{array}{ll} {\epsilon} ( y - \hat{y}) 
& \mbox{if } y - \hat{y} \geq 0 \\
(1-{\epsilon}) ( \hat{y} -y) & \mbox{if } y - \hat{y} < 0 \end{array} \right.
\end{displaymath}
Suppose $\hat{q}_{\epsilon} (\cdot)$ represents
the estimated conditional quantile function.  In Section~5 we describe
our approach of using normalized conformal prediction as a starting point
for quantile regression.  Quantile regression does not promise marginal
coverage, so the initial interval endpoints, $\hat{q}_{0.5 \alpha} (x)$ and 
$\hat{q}_{1 - 0.5 \alpha} (x)$, need modification.
We apply the CQR\cite{rpc19} technique, which uses the nonconformity score
\begin{displaymath}
\eta_Q (z^i) \; = \; \max \{ \hat{q}_{0.5 \alpha} (x^i) - y^i, \;
y^i - \hat{q}_{1- 0.5 \alpha}(x^i) \}.
\end{displaymath}
Let $\varepsilon_1 , \ \varepsilon_2 , \ \dots , \ 
\varepsilon_{| {\cal Z}_C |}$ be the sorted nonconformity measures of each 
example in the calibration set in nonincreasing order.  Then for a new input 
image $X^j = x^j$, the prediction interval
\begin{displaymath}
[\hat{q}_{0.5 \alpha} (x^j) - \varepsilon_m , \;
\hat{q}_{1-0.5 \alpha} (x^j) + \varepsilon_m ]
\end{displaymath}
offers valid marginal coverage.  

\section{On EDGENet and the Simulated Dataset}
The original EDGENet\cite{csy2019} predicts a $2 \times 1024$ matrix of
left and right line edge positions from a $64\times1024$ noisy simulated 
SEM image of one rough line with pixel size $0.5 \times 2$ nm and an unknown 
level of Poisson noise; see Figure~1. The edge positions are output
with pixel-level precision and not with subpixel-level precision due to
the simulator used to generate the data for References~29, 30, and 42. 
EDGENet employs seventeen convolutional 
layers \cite{convolutional} with filter dimension $3 \times 3 \times 
\mbox{input depth}$.  The first four convolutional layers each employ 
64 filters, the next four each employ 128 filters, the following four each 
employ 256 filters and the subsequent four each employ 512 filters.  
These sixteen convolutional layers are each followed by a batch normalization 
layer \cite{batch} and a dropout layer \cite{dropout} with dropout probability
 of 0.2 for regularization. The last convolutional layer employs one filter 
to output the $2 \times 1024$ matrix of estimated edge positions. 
Figure~2 illustrates the sizes of the output volumes or tensors associated
with the convolutional layers. 
The mean absolute error (MAE) is the training loss criterion.
For this work we augmented the original EDGENet architecture with one extra 
layer to compute the LER values associated with the left and right edges.  

\begin{figure}
\begin{center} 
\includegraphics[scale=0.5]{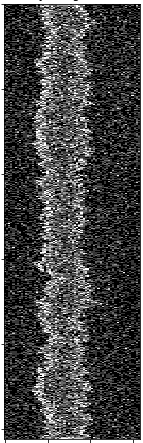}
\caption{A noisy SEM image consisting of one line with two edges
of dimension $64 \times 1024$. The aspect ratio of the image has been 
altered to facilitate viewing.  Reprinted with permission from Ref.~46.}   
\end{center}
\label{fig:line}
\end{figure} 
\begin{figure}
\begin{center} 
\includegraphics[scale=0.5]{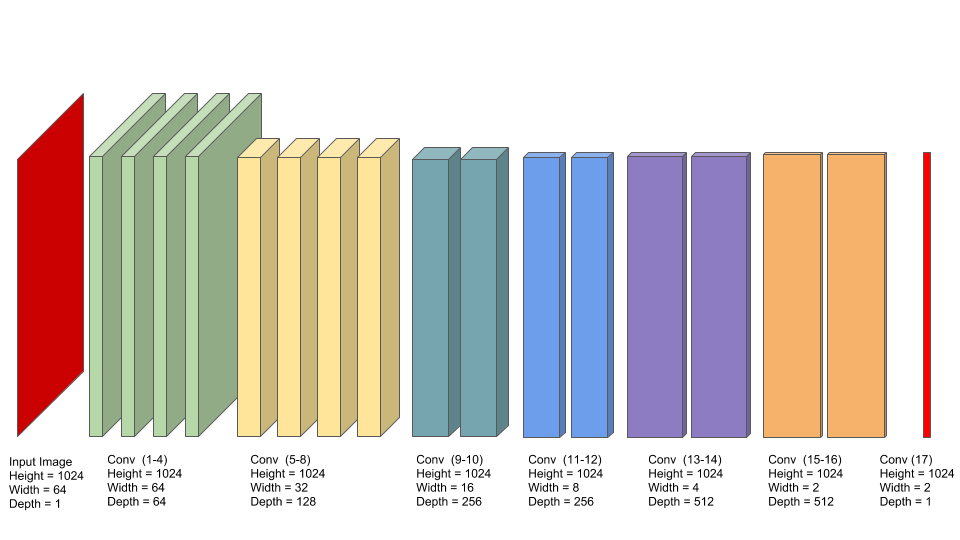}
\caption{ The 17 convolutional layers of EDGENet.  
Reprinted with permission from Ref.~30.}   
\end{center}
\end{figure} 

Because of the popularity of simulation within semiconductor metrology,
our dataset consists of the simulated SEM images used earlier for
References~29, 30 and 42.  The generation of the dataset involves multiple
steps.  First, apply the Thorsos method\cite{thorsos, mack2013a} with normally
distributed random variables to simulate rough line edges or linescans
which each follow a Palasantzas spectral model\cite{palasantzas} 
characterized by three parameters: 
$\sigma$ is the line edge roughness (LER), i.e., the standard deviation of 
edge positions, $\alpha$ is the roughness (or Hurst) exponent and 
$\xi$ is the correlation length:
$$PSD(f) = \frac{\sqrt{\pi}\Gamma(\alpha + 0.5)}{\Gamma(\alpha)} . \frac{2\sigma^2 \xi}{(1 + (2\pi f\xi)^2)^{\alpha + 0.5}}. $$

Each simulated edge has length 2.048 microns or, equivalently, 1024 pixels.
Our edges have eight possible LER values ($\sigma$ = 0.4, 0.6, 0.8, 
1.0, 1.2, 1.4, 1.6, 1.8 nm), nine possible Hurst/roughness exponent values 
($\alpha$ = 0.1, 0.2, 0.3, 0.4, 0.5, 0.6, 0.7, 0.8, 0.9) and 35 possible
correlation length values ($\xi$ = 6, 7, ..., 40 nm) for a total of 
2520 possible combinations of parameters ($\sigma$, $\alpha$, $\xi$).
For each combination we produced eight edges.

The next step in generating the dataset is to apply a SEM simulator to
generate the rough-line images from the simulated rough edges.
A Monte Carlo-based simulator like JMONSEL\cite{jmonsel} offers the
most realistic simulated images, but we opt in References~29, 30 and 42
for the much faster simulator 
ARTIMAGEN\cite{cizmar2008simulated,cizmar2009optimization} to generate
a large dataset and describe there the parameter settings we use; both
simulators are publicly available.
The images from ARTIMAGEN ``closely mimic the noise, contrast and resolution,
drift, vibration'' of real SEM images\cite{pv2011}.
Our 10080 images have dimension $64 \times 1024$ pixels with 
pixel width $0.5$ nm and pixel height $2$ nm and incorporate
a rough line at varying locations within the image of width 10 nm or 15 nm 
with two of the previously generated edges together with  random 
backgrounds, a fixed edge effect, fine structure and Gaussian blur.
The ARTIMAGEN library does not offer fractional edge positions, so the edge 
positions are rounded; this is a limitation of ARTIMAGEN.  Rounding 
operations are not differentiable, so we do not use them in the implementation
in the new final layer of EDGENet nor in the actual LER computations. 

We have so far described our ``original'' image set.  We apply
a feature of the ARTIMAGEN library to produce ten images corrupted by
varying levels of Poisson noise for each original image with 
electron density per pixel in the range 
$\{2, \ 3, \ 4, \ 5, \ 10, \ 20, \ 30, \ 50, \ 100, \ 200\}$.  These
100800 images are our noisy image dataset. Our supervised learning dataset 
for the training of the original EDGENet consists of pairs of matrices 
$(x^i, \ y^i)$, where $x^i$ is a noisy image and $y^i$ is a $2 \times 1024$ 
matrix of edge positions associated with
corresponding original image.  For the augmented EDGENet 
$y^i$ is now either the LER of the left edge of the original
image or the LER of the right edge of the original image.

\section{Auxiliary Neural Networks}

Our first attempts\cite{b6} at using conformal prediction and quantile
regression were preliminary and did not take advantage of our group's 
earlier work on denoising.  However, we now seek to use the relationship
between the amount of noise in an image and the inaccuracies in an edge
detection procedure for that image.  The noise image presumably contains
information to help a neural network predict a function of the residual,
namely, $- \ln | y^i - g(x^i)|.$  Our results support this intuition.

We propose three similar deep neural networks NCPMaxPoolNet, NCPLSTMNet, 
and NCPBLSTMNet that each improve upon the average interval lengths 
that we reported in Ref.~23. All of them begin with a CNN that learns 
multiple filter maps from the noise image and subsequently employ an
individual feature extraction layer followed by a fully connected network 
that predicts the desired function of the residual. Each CNN contains a 
convolution layer of 64 filters followed by another with 128 filters
followed by a third convolutional layer with 256 filters. All layers
employ a 3$\times$3 kernel, 2$\times$2 stride length and equal padding. 
We initialize the convolution weights via Xavier uniform initialization and 
use the Rectified Linear Unit (ReLU) activation function. Each convolutional 
layer is followed by a batch normalization layer and a dropout layer 
with a 2\% dropout rate. The output of this portion of each network
is 256 feature maps of dimensions 64$\times$128. 

The next set of layers reduce the dimensionality of the output filter maps
and extract features to facilitate estimation. NCPMaxPoolNet uses a global 
max-pooling layer to reduce the $64 \times 128 \times 256$ dimensional output 
from the CNN to a $1 \times 1 \times 256$ vector.  The NCPLSTMNet and 
NCPBLSTMNet networks follow the initial CNN layers with another convolution 
layer with 64 filters to produce 64 feature maps of dimension 32$\times$64 
and subsequently apply a 4$\times$4 max-pooling layer to output feature maps 
of dimension 8$\times$16. The NCPLSTMNet output is reshaped into 64 vectors
of length 128 and input to a Long Short Term Memory (LSTM) \cite{b18} layer
with 64 cell units to produce a 64-dimensional vector.  Since
Bidirectional Long Short Term Memory (BLSTM) networks often outperform
LSTM networks\cite{b20} the NCPBLSTMNet variant replaces the LSTM layer 
with a BLSTM layer.

All three network models end with a set of fully connected layers. 
NCPMaxPoolNet is designed to predict error estimates for both the 
left and right edges and consists of two branches of four fully connected 
layers with sixty-four, sixteen, four, and one neuron, respectively, each
using the ReLU activation function.   As reported in our earlier work 
\cite{b6}, the ARTIMAGEN simulation process yields systematic differences 
in the left and right edge data mainly because of the application of Gaussian 
blur at $30^{\circ}$ to the horizontal x-axis, which results in random drift 
effects typically known as astigmatic effects.   Therefore,
one can construct more efficient prediction intervals by training
separate models for each edge.  For NCPLSTMNet and NCPBLSTMNet we use
this strategy and use four fully connected layers with thirty-two, eight, 
four, and one neuron, respectively; the first three layers apply a
Leaky ReLU \cite{b21} activation function and the fourth one applies a
ReLU activation function. We use the mean absolute error loss function
to train all three networks.  As we will see, these strategies offer
performance improvements over the basic version of conformal prediction.

Just as effective modeling is needed to create efficient normalized
conformal prediction intervals, it is not immediately apparent how to
effectively use quantile regression on high-dimensional image data.
Our approach takes advantage of the strengths of our new normalized
conformal prediction models together with the success of residual 
learning \cite{he2016deep} and a variation of stacked generalization 
\cite{wolpert}.  We propose three quantile regression neural networks
QRMaxpoolNet, QRMaxpool-LSTMNet, and QRMaxpool-BLSTMNet and train separate
models for the left and right edges. Each network inputs the EDGENet's
LER prediction along with estimates $\phi (x_N^i )$ of the error measure
$- \ln | y^i - g(x^i)|$ associated with the desired miscoverage rate
from the normalized conformal prediction model.
QRMaxpoolNet inputs the EDGENet prediction and the estimated error measure
from NCPMaxPoolNet.  QRMaxpool-LSTMNet and QRMaxpool-BLSTMNet each take
three inputs:  the EDGENet prediction, the estimated error measure
from NCPMaxPoolNet and either the estimated error measure from
NCPLSTMNet or from NCPBLSTMNet.  As depicted in Fig.~3, each
network has one hidden layer and an output layer.  The hidden layer is 
fully connected with a number of neurons equal to twice the number of 
network inputs.  The output layer estimates the upper and lower quantile
levels.  The network architecture incorporates a residual connection between
the EDGENet LER input and the output layer so that the model alters the
LER prediction to output the endpoints of a prediction interval.
These neural networks are trained using the pinball loss function \cite{kb78},
but the asymmetric Huber may offer more flexibility in meeting business
goals \cite{uber}.
Applying CQR to each of these networks yields models we call
CQRMaxpoolNet, CQRMaxpool-LSTMNet, and CQRMaxpool-BLSTMNet.

\begin{figure}[thbp]
\centering
\includegraphics[width=0.85\textwidth]{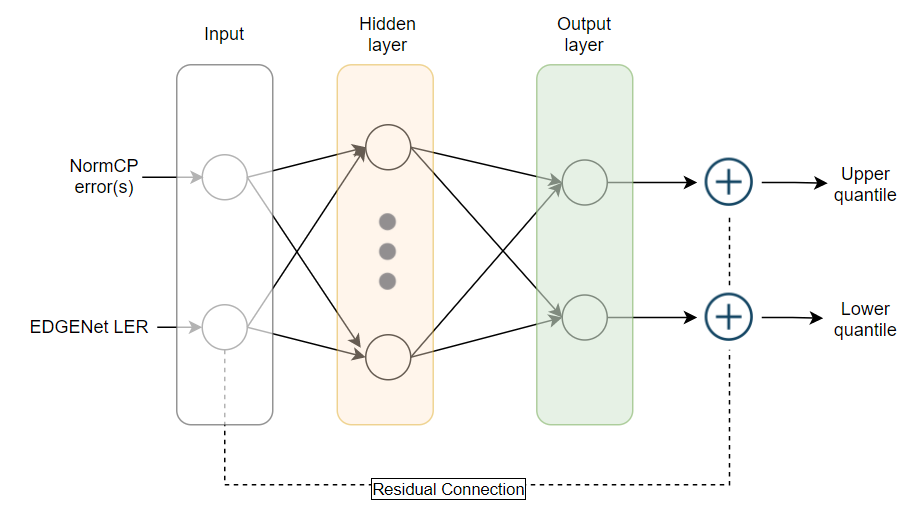}
\caption{Quantile regression neural network architecture.}
\label{fig3}
\end{figure}

\section{Experiments and Results}
We now examine the results of the proposed schemes and present
the coverage and interval sizes for both the left and right edges at a 
10\% miscoverage rate.  As we mentioned earlier, there are systematic 
differences between the left and right edge data, so we gain more
insight about the efficacy of the proposed techniques.

The six proposed neural network models were trained on an Intel Xeon E5-2680 
v4 processor running at 2.4 GHz and a Tesla K80 GPU.  We use hyperparameter 
optimization to choose values for the batch size, optimizers, and the 
learning rate and determine that a batch size of eighteen combined with 
an Adam\cite{kingma2014adam} optimizer with a learning rate of 0.001 offers 
the best training performance over all models.

The 100800 noisy images are first normalized so that all values are in 
the range (0,1).  We apply SEMNet to each noisy image to obtain its 
corresponding noise image.  The noise image dataset is separated into 
a proper training set, a calibration set, and a test set. 
The combination of calibration and test sets consist of the 11520 noise images
associated with correlation length $\xi$ in the set 
$\{10, \ 20, \ 30, \ 40 \}$~nm which are partitioned into two sets of 5760 
images for the calibration and test sets based on the corresponding
original images to approximate exchangeability\cite{b6}; for each original
image all ten corresponding noise images all belong to the same set. 
For the quantile regression neural networks, we calculate EDGENet's LER 
predictions and the associated normalized conformal prediction error 
measures for each scheme for the entire dataset as part of the 
quantile regression procedure.

We present the coverage and interval statistics for the left and right edges
at 10\% miscoverage rate in Tables~1 and 2, respectively.  Nearly all of
the proposed conformal models outperform the conformal prediction and
normalized conformal prediction schemes of Ref.~23, and in some cases
the performance improvement is significant.  The proposed 
normalized conformal prediction schemes perform particularly well
on the left edge data. Quantile regression networks have been investigated
for multiple applications and are known to have a tendency to undercover 
\cite{tlp18};  our quantile regression networks grossly undercover,
but conformalization helps to address this problem even though the relative
size of our calibration set is much lower than the recommendation of
Ref.~41.  In Ref.~23, we reported that EDGENet has more difficulty with 
prediction for right edge data compared with left edge data.  Despite this,
the proposed conformalized quantile regression models reduce the disparity
between average interval lengths for the left edge and right edge data.

\begin{table}[htbp]
\caption{Coverage and interval width summary statistics for left edge LER at 10\% miscoverage rate}
\begin{center}
\begin{tabular}{|c|c|c|}
\hline
\textbf{Method} & \textbf{Coverage} & \textbf{Average interval} \\

\textbf{} & \textbf{(\%)} & \textbf{length (nm)} \\
\hline
Conformal Prediction (CP)\cite{b6} & 90.22 & 0.135 \\
\hline
Normalized CP\cite{b6} & 90.15 & 0.153 \\
\hline
NCPMaxPoolNet & 89.36 & 0.127 \\
\hline
NCPLSTMNet & 90.30 & 0.130 \\
\hline
NCPBLSTMNet & 90.43 & 0.124 \\
\hline
QRMaxpoolNet & 81.39 & 0.146 \\
\hline
CQRMaxpoolNet & 87.97 & 0.169 \\
\hline
QRMaxpool-LSTMNet & 71.39 & 0.094 \\
\hline
CQRMaxpool-LSTMNet & 88.18 & 0.134 \\
\hline
QRMaxpool-BLSTMNet & 79.60 & 0.096 \\
\hline
CQRMaxpool-BLSTMNet & 88.14 & 0.117 \\
\hline
\end{tabular}
\label{tab1}
\end{center}
\end{table}

\begin{table}[htbp]
\caption{Coverage and interval width summary statistics for right edge LER at 10\% miscoverage rate}
\begin{center}
\begin{tabular}{|c|c|c|}
\hline
\textbf{Method} & \textbf{Coverage} & \textbf{Average interval} \\

\textbf{} & \textbf{(\%)} & \textbf{length (nm)} \\
\hline
Conformal Prediction (CP) \cite{b6} & 89.22 & 0.186 \\
\hline
Normalized CP \cite{b6} & 88.96 & 0.241 \\
\hline
NCPMaxPoolNet & 89.58 & 0.242 \\
\hline
NCPLSTMNet & 89.22 & 0.186 \\
\hline
NCPBLSTMNet & 89.24 & 0.181 \\
\hline
QRMaxpoolNet & 84.83 & 0.156 \\
\hline
CQRMaxpoolNet & 88.58 & 0.168 \\
\hline
QRMaxpool-LSTMNet & 65.59 & 0.077 \\
\hline
CQRMaxpool-LSTMNet & 89.01 & 0.128 \\
\hline
QRMaxpool-BLSTMNet & 75.64 & 0.096 \\
\hline
CQRMaxpool-BLSTMNet & 89.35 & 0.133 \\
\hline
\end{tabular}
\label{tab2}
\end{center}
\end{table}

\section{Ongoing Work and Conclusions}
We are aware of the numerous and rapid advances related to attention
mechanisms, and we are working to leverage attention in our distribution-free
prediction interval models.

The semiconductor manufacturing community will continue to impact
computing.  Those who may benefit from continued advances in computing 
must help them develop 
more confidence in the role of deep learning and other forms of artificial 
intelligence for decision making.  Distribution-free prediction intervals 
with coverage guarantees are an initial step to address this need, and 
much more research is required.

\acknowledgments 
The authors used the Texas A\&M University High Performance Research Computing
Facility to conduct part of the research.

\vspace{1ex}

\listoffigures
\listoftables

\end{spacing}
\end{document}